\PassOptionsToPackage{colorlinks=true,urlcolor=blue}{hyperref}
\documentclass{article}

\usepackage{arxiv}

\usepackage[utf8]{inputenc} 
\usepackage[T1]{fontenc}    

\usepackage{url}            
\usepackage{booktabs}       
\usepackage{array}          
\usepackage{hhline}         
\usepackage{amsfonts}       
\usepackage{amsmath}        
\usepackage{siunitx} 
\usepackage[table]{xcolor} 
\usepackage{nicefrac}       
\usepackage{microtype}      
\usepackage{lipsum}		
\usepackage{graphicx}
\usepackage{natbib}
\usepackage{doi}

\usepackage{cleveref}
\usepackage{booktabs}
\usepackage{tabularray}
 \usepackage{multirow}
\usepackage[colorlinks=true,      
            urlcolor=blue]{hyperref}  
\title{VoRTeC: Taming Foundation Flow for One-step Real time Video Compression}

\author{
    {\hspace{1mm}Yichong Xia}\thanks{These authors contributed equally to this work.}\\
    Shenzhen International Graduate School\\
    Tsinghua University\\
    \texttt{xiayc23@tsinghua.edu.cn}\\
    \And
    {\hspace{1mm}Shiyu Qin}\footnotemark[1]\\
    Shenzhen International Graduate School\\
    Tsinghua University\\
    \texttt{qinsy23@mails.tsinghua.edu.cn}\\
    \And
    {\hspace{1mm}Qinhong Wu}\\
    Harbin Institute of Technology, Shenzhen\\
    \texttt{2024111108@stu.hit.edu.cn}\\
    \And
    {\hspace{1mm}Bin Chen}\thanks{Corresponding author.}\\
    Department of Computer Science\\
    Harbin Institute of Technology, Shenzhen\\
    \texttt{chenbin2021@hit.edu.cn}\\
    \And
    {\hspace{1mm}Zeyuan Chen}\\
    School Of Computer Science\\
    Peking University\\
    \texttt{chenzy01@stu.pku.edu.cn}\\
    \And
    {\hspace{1mm}Haoqian Wang}\\
    Shenzhen International Graduate School\\
    Tsinghua University\\
    \texttt{wanghaoqian@tsinghua.edu.cn}\\
}

\renewcommand{\shorttitle}{\textit{arXiv} Template}

\hypersetup{
pdftitle={A template for the arxiv style},
pdfsubject={q-bio.NC, q-bio.QM},
pdfauthor={David S.~Hippocampus, Elias D.~Striatum},
pdfkeywords={First keyword, Second keyword, More},
}

\begin{document}
\maketitle

\begin{abstract}

    Ultra-low bitrate video compression still faces critical challenges: traditional neural video compression inevitably introduces blurring artifacts, while diffusion-based generative video compression suffers from excessive decoding latency and poor temporal consistency. To address these issues, we propose $\mathtt{VoRTeC}$, a Video Compression framework built upon a foundational flow model (Wan2.1). By compactly encoding latent video representations, predicting the positions of compressed representations along flow trajectories, and integrating multi-scale priors, $\mathtt{VoRTeC}$ enables the compressor to harness generative video flow priors effectively. Without accessing the parameters or gradients of flow matching networks, our framework achieves one-step decoding and reconstructions with high perceptual fidelity. Meanwhile, we maintain consistency across frame groups via tail-frame reuse and prior caching. Extensive experiments demonstrate that our method reduces bit consumption by 58\% compared to prior diffusion-based approaches, with decoding speed boosted by 3 to 197 times: $\mathtt{VoRTeC}$ achieves a decoding speed of 13 FPS at 720p and 32 FPS at 480p. \href{https://darc8-sun.github.io/VoRTec_compress/}{Project page: https://darc8-sun.github.io/VoRTec\_compress/}.

\end{abstract}

\section{Introduction}
\label{sec:introduction}

Video compression underpins virtually every modern visual communication pipeline, from real-time streaming and video conferencing to cloud storage and autonomous driving~\citep{sullivan2012hevc,bross2021vvc}.
As bandwidth demands continue to escalate, the ability to reconstruct perceptually faithful video at extremely low bitrates has become a critical yet unresolved challenge.

\begin{figure}[t!]
\centering
\includegraphics[width=1.0\linewidth]{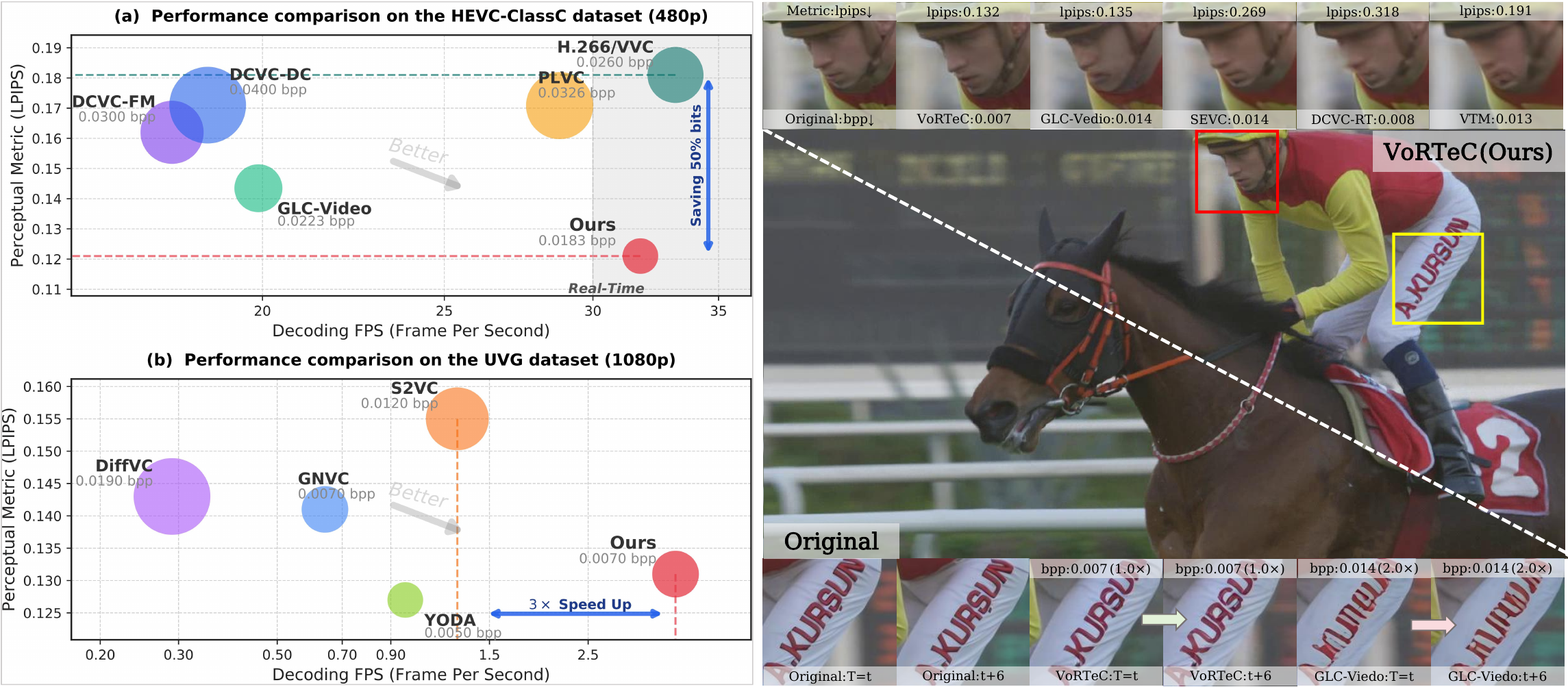}
\caption{Left: Comparison of rate-perception performance and decoding efficiency. $\mathtt{VoRTeC}$ (Ours) delivers a better rate–perception trade-off than prior diffusion-based video codecs and enables real-time decoding, running up to $3\times$ faster than existing diffusion-based methods. Right: Qualitative comparison between baseline and our proposed approach. $\mathtt{VoRTeC}$ (Ours) is capable of reconstructing complex textures with realism, even at extremely low bit rates. \emph{Best viewed when zoomed in.}}
\label{fig:firstpic}
\end{figure}

Traditional hybrid codecs such as HEVC~\citep{sullivan2012hevc} and VVC~\citep{bross2021vvc} rely on block-wise prediction and transform coding, which produce visible blocking and ringing artifacts under aggressive quantization.
Recent neural video codecs (NVCs)~\citep{lu2019dvc,li2021dcvc,li2023dcvc,li2024dcvcfm,jia2025dcvcrt} have surpassed these standards in rate--distortion optimization, yet distortion-driven objectives (e.g., MSE) inherently oversmooth textures and erase fine structures when the bitrate drops to the ultra-low regime, causing a sharp decline in perceptual realism.

To improve perceptual quality at low bitrates, a growing body of work introduces generative priors into the compression pipeline---a paradigm known as generative compression~\citep{mentzer2022hific,careil2023perco,vonderfecht2025diffc}.
In the image domain, large-scale GANs~\citep{goodfellow2014gan} and diffusion models~\citep{ho2020ddpm,rombach2022ldm} have been successfully leveraged to recover high-frequency details, yielding visually convincing reconstructions even below 0.01~bpp.
Extending this paradigm to video, however, introduces a fundamentally stricter requirement: \emph{temporal coherence}.
Early perceptual video codecs~\citep{mentzer2022nvcdisc,yang2022plvc} incorporate adversarial training or perceptual losses, but their limited generation capacity still yields noticeable artifacts.
More recent diffusion-based video codecs~\citep{ma2025diffvc,qi2025glcvideo} adopt pre-trained \emph{image} diffusion priors for frame-wise enhancement.
While each frame may appear sharp in isolation, image-domain priors lack any modeling of temporal dynamics, causing restored textures to drift across frames---a phenomenon known as \emph{perceptual flickering}---which becomes especially severe at ultra-low bitrates.

Video-native diffusion models, particularly diffusion transformers (DiTs)~\citep{peebles2023dit,wan2025,yang2024cogvideox}, offer a natural remedy.
Trained on large-scale video data, they learn spatio-temporal latent representations that capture appearance, motion, and long-range dependencies within a unified structure.
Very recently, GNVC-VD~\citep{mao2025gnvcvd} demonstrated that integrating a pre-trained video diffusion transformer (Wan2.1~\citep{wan2025}) into neural video compression can produce coherent fine textures with strong temporal stability.
However, this benefit comes at a prohibitive cost: the decoder must perform \emph{multi-step} flow-matching inference---typically 5 or more sequential denoising iterations through a multi-billion-parameter DiT.
Similarly, Free-GVC~\citep{ling2026freegvc} reformulates compression as progressive coding along the diffusion trajectory, but its multi-step reverse process yields less than 0.5~fps even at 480p.
In essence, existing video-native generative codecs face a fundamental \emph{speed--quality dilemma}: the multi-step sampling process that endows video diffusion priors with their generative power is precisely what makes them impractical for real-time decoding.

A parallel line of research addresses diffusion efficiency through distillation.
Distribution Matching Distillation (DMD)~\citep{yin2024dmd,yin2024dmd2} and consistency models~\citep{song2023consistency} enable single-step generation in the image domain, and S2VC~\citep{xue2025s2vc} has recently applied single-step image diffusion to video compression.
Yet these methods remain anchored to \emph{image} diffusion backbones, which lack the spatio-temporal modeling required for coherent video reconstruction.
The key question is therefore: \emph{can we fully use the rich spatio-temporal prior of a video-native flow-matching foundation model into a single decoding step, while preserving both perceptual fidelity and inter-frame consistency?}

We propose $\mathtt{VoRTeC}$, a \textbf{R}eal \textbf{T}ime \textbf{V}ideo \textbf{C}ompr\textbf{e}ssion framework built upon
a \textbf{o}ne-step foundational Flow model, as one solution to this problem. $\mathtt{VoRTeC}$ compresses spatiotemporal latents into compact transmittable representations, and treats the decoding result as a temporal state along the flow path. This enables us to perform enhanced decoding even without access to the parameters or gradients of the foundation flow-matching network. Then, through a ViT-based prior fusion network, $\mathtt{VoRTeC}$ fuses and aligns the coarse-grained compressed representation with the fine-grained prior representation, ultimately achieving fast single-step decoding. Furthermore, we design an inter-group communication technique (CGG), which allows spatiotemporal information from preceding video groups to flow into subsequent groups. This ensures inter-group consistency and further reduces the bitrate without incurring additional training or decoding overhead. We conduct extensive experiments at various resolutions to validate the efficiency of $\mathtt{VoRTeC}$: $\mathtt{VoRTeC}$ achieves a decoding speed of \emph{3.93 fps on 1080p} datasets and enables \textbf{real-time} decoding at \emph{32 fps on 480p}. Compared with previous diffusion-based baselines, our method achieves drastically reduced decoding latency while improving perceptual quality by approximately 58.12\%–73.25\%, as illustrated in \Cref{fig:firstpic}.
Furthermore, due to its externally mounted architectural design, $\mathtt{VoRTeC}$ exhibits low training complexity; consequently, even when employing Wan2.1 \citep{wan2025wan} as the backbone, it can be effectively trained on a single NVIDIA A6000 GPU.

Building upon this, we further propose $\mathtt{VoRTeC}^{+}$, which fine-tunes the flow-matching network with LoRA on top of $\mathtt{VoRTeC}$. Benefiting from the excellent prior pattern learning paradigm of $\mathtt{VoRTeC}$, $\mathtt{VoRTeC}^{+}$ improves perceptual performance by 20–30\% over the original scheme, while introducing only 10M additional parameters and leaving the inference speed unaffected, surpassing all existing video compression approaches built upon foundation flow models. As shown in the \Cref{fig:firstpic}, our proposed method can reconstruct finer and more realistic details, and significantly outperforms the prior generative method GLC‑Video \citep{qi2025glcvideo} in terms of temporal consistency.

In summary, our contributions can be summarized as follows:

\begin{itemize}

   \item We propose a latent codec that re-compresses spatiotemporal latent representations into compact, quantization-ready codes and employ it to initialize the flow-matching process by mapping these compressed representations onto the corresponding flow trajectories. This strategy enables the framework to capture the underlying prior distribution even when the initial gradients of the flow-matching network are unavailable.

   \item We design the Flow Prior Multi-Fusion module to perform multi-scale fusion and alignment between the prior and compressed latent. On one hand, this enables the compressor to better remove prior redundancy; on the other hand, it further refines the prior representation. Meanwhile, we propose an inter-group communication algorithm that ensures temporal consistency across frame groups in a training-free manner.

   \item Building upon these insights, we propose $\mathtt{VoRTeC}$ and $\mathtt{VoRTeC}^{+}$. Through solely external module and algorithm design, $\mathtt{VoRTeC}$ learns the prior distribution of the foundation flow and achieves high-fidelity single-step decoding. Compared to diffusion-based video compression baselines, $\mathtt{VoRTeC}$ achieves BD-rate savings of 58.12\%–73.25\% in terms of LPIPS. Through further LoRA fine-tuning, $\mathtt{VoRTeC}^{+}$ extends this performance advantage to 64.53\%–74.8\%. Compared to previous Wan2.1 based method \cite{mao2025gnvcvd}, $\mathtt{VoRTeC}^{+}$ achieves a 23.44\% improvement in perceptual quality (DISTS). Meanwhile, $\mathtt{VoRTeC}$ substantially improves the decoding speed over previous diffusion-based methods ($3\sim 197\times$), achieving a decoding frame rate of 32 fps on 480p and requiring only 0.25 s to decode a single 1080p frame.
\end{itemize}

\section{Related Work}
\label{sec:related_work}

\subsection{Neural Video Compression}
Neural video compression (NVC) has advanced rapidly, leveraging end-to-end learning to surpass traditional hybrid codecs in performance.
The fundamental objective of lossy compression is the rate-distortion optimization:
\begin{equation}
\label{eq:rd}
\min_{\boldsymbol{\theta}}\; R\!\left(\hat{\mathbf{x}};\, \boldsymbol{\theta}\right) + \lambda\, D\!\left(\mathbf{x},\, \hat{\mathbf{x}}\right),
\end{equation}
where $R(\hat{\mathbf{x}}; \boldsymbol{\theta})$ estimates the bitrate of the compressed representation $\hat{\mathbf{x}}$, $D(\cdot, \cdot)$ is a distortion metric (e.g., MSE), and $\lambda$ controls the rate--distortion trade-off.
DVC~\citep{lu2019dvc} introduced the first end-to-end framework that jointly optimizes motion estimation, residual coding, and entropy modeling.
DCVC~\citep{li2021dcvc} shifted the paradigm from residual coding to conditional coding, allowing the model to learn more informative temporal context from decoded features.
Subsequent extensions progressively enhanced context modeling: DCVC-FM~\citep{li2024dcvcfm} incorporated feature modulation for finer rate control, and DCVC-RT~\citep{jia2025dcvcrt} optimized both architecture and runtime efficiency to achieve real-time decoding on consumer hardware.
Other lines of work explore hierarchical variational autoencoders~\citep{lu2024dhvc} and non-local temporal correlations~\citep{jiang2025ecvc}.
Despite these advances, distortion-oriented NVCs optimized for MSE or MS-SSIM~\citep{wang2003msssim} inevitably produce over-smoothed reconstructions at ultra-low bitrates, lacking fine textures and perceptual realism.

\subsection{Generative Compression}
Generative compression leverages learned generative models to recover realistic textures beyond pixel-level fidelity, making it particularly effective at ultra-low bitrates.
The shift from classical R--D optimization to the rate--distortion--perception (R--D--P) trade-off~\citep{blau2019rdp} can be formalized as:
\begin{equation}
\label{eq:rdp}
\min_{\boldsymbol{\theta}}\; R\!\left(\hat{\mathbf{x}};\, \boldsymbol{\theta}\right) + \lambda\, D\!\left(\mathbf{x},\, \hat{\mathbf{x}}\right) + \gamma\, d\!\left(p_{\mathbf{x}},\, p_{\hat{\mathbf{x}}}\right),
\end{equation}
where $d(\cdot, \cdot)$ measures the distributional discrepancy between the real and reconstructed data distributions $p_{\mathbf{x}}$ and $p_{\hat{\mathbf{x}}}$, and $\gamma$ controls the emphasis on perceptual realism.

In the image domain, HiFiC~\citep{mentzer2022hific} pioneered perceptual and adversarial losses with a generative decoder, while subsequent work~\citep{li2025diffeic,xia2025diffpc,xia2026diffcr,jia2026codlite} increasingly turned to diffusion models as stronger priors that guide multi-step denoising with compact semantic latents.
To address sampling latency, StableCodec~\citep{zhang2025stablecodec} and OneDC~\citep{xue2025onedc} distill the diffusion process into a single step, achieving over $20\times$ speedup.
These advances, however, are inherently limited to single-frame reconstruction without temporal modeling.
Extending generative compression to video, GLC-Video~\citep{qi2025glcvideo} and DiffVC~\citep{ma2025diffvc} adopt image-domain priors for frame-wise enhancement but inevitably suffer from temporal flickering due to the absence of inter-frame modeling.
GNVC-VD~\citep{mao2025gnvcvd} introduced the first video-native diffusion prior, achieving temporally coherent refinement via multi-step flow matching, while S2VC~\citep{xue2025s2vc} pursued faster single-step image-diffusion decoding at the cost of inter-frame consistency.

\subsection{Video Diffusion Models and Flow Matching}
Diffusion models~\citep{ho2020ddpm} have advanced video generation from 2D U-Nets~\citep{ho2022vdm} through latent diffusion~\citep{rombach2022ldm} and video latent models~\citep{yang2024cogvideox,blattmann2023svd} to the diffusion transformer (DiT)~\citep{peebles2023dit}, which models videos as token sequences for long-range temporal reasoning.
Wan2.1~\citep{wan2025} further advanced this paradigm with a spatio-temporal VAE and a flow-matching~\citep{lipman2023flowmatching,liu2023rectified} formulation.
Unlike DDPM's discrete Markov chain, flow matching learns a continuous velocity field trained via:
\begin{equation}
\label{eq:fm_loss}
\mathcal{L}_{\text{FM}} = \mathbb{E}_{t,\, \mathbf{x}_0,\, \mathbf{x}_1}\left[\left\| \mathbf{v}_\theta(\mathbf{x}_t, t) - (\mathbf{x}_1 - \mathbf{x}_0) \right\|^2\right],
\end{equation}
where $\mathbf{x}_t = (1{-}t)\,\mathbf{x}_0 + t\,\mathbf{x}_1$, $t \in [0,1]$, and $\mathbf{v}_\theta$ is a DiT-based velocity network.


\begin{figure*}[t]
\centering
\includegraphics[width=1\linewidth]{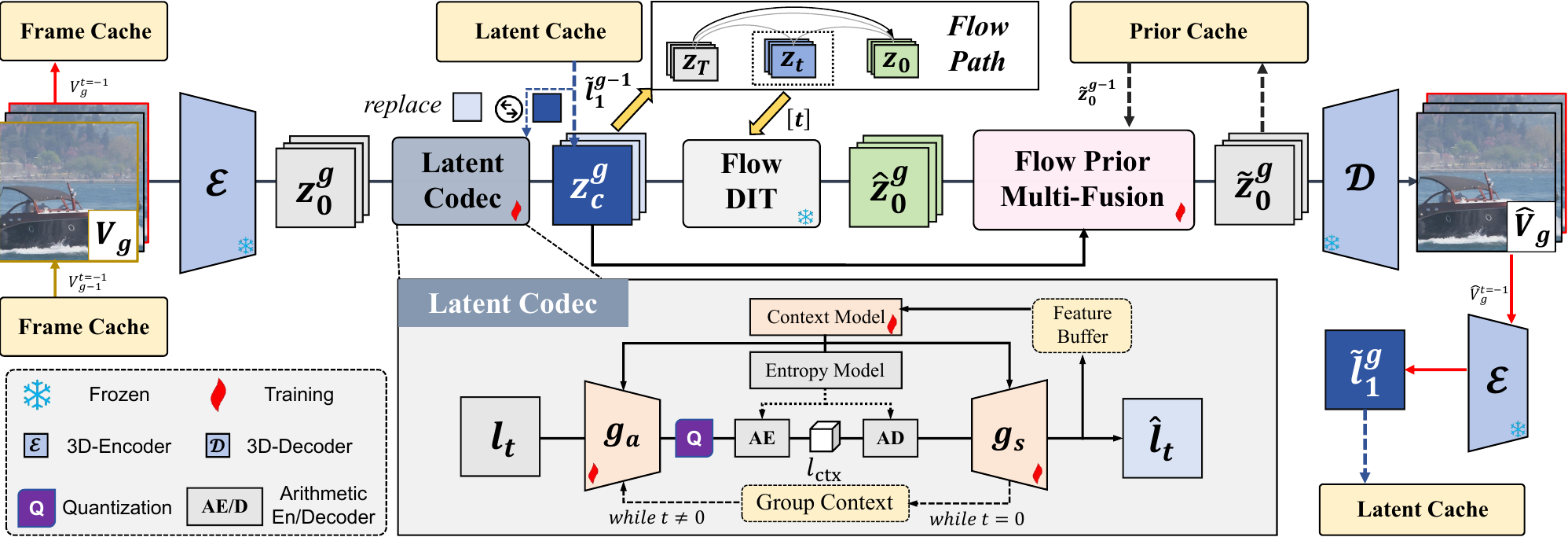}
\caption{Overall pipeline of $\mathtt{VoRTeC}$. \(\mathcal{E}\) and \(\mathcal{D}\) denote the pre-trained 3D video en/decoder \citep{wan2025}, and FlowDIT adopts the flow-matching network from Wan2.1 (1.3B) \cite{wan2025}.}
\label{fig:pipeline}
\end{figure*}

\section{Methodology}

\subsection{Framework Overview}
\label{subsec:Framework}
The framework of our proposed $\mathtt{VoRTeC}$ is illustrated in \Cref{fig:pipeline}: given an input group of frames $\boldsymbol{V}_g \in \mathbb{R}^{(1+T) \times H \times W \times 3}$, a 3D analysis encoder $\mathcal{E}(\cdot)$ compresses it into a compact spatiotemporal latent representation:
\begin{align}
\{l^g_t\}^{1+T/4}_{t=1}=z^g_0=\mathcal{E}(\boldsymbol{V}_g)
\end{align}
$\mathcal{E}(\cdot)$ performs temporal modeling through causal convolution: it encodes the first frame into an Intra-latent (I-latent) $l^g_0$, and applies temporal downsampling to the subsequent frames, encoding them as Predictive latents (P-latents) $l^g_{t\geq 1}$. This ensures spatiotemporal consistency within each frame group, but continuity across frame groups cannot be guaranteed. Inspired by \citep{sullivan2012hevc}, we partition the entire video sequence into key groups (also referred to as I-Groups) and predictive groups (P-Groups).
\subsubsection{Key Group Compression}
We refer to the first frame group of a meta-group as the key group. To reduce intra-group redundancy, our latent codec needs to model latent-wise context. Specifically:
\begin{align}
\hat{y}^g_t=\mathbf{Q}\left(\boldsymbol{g}_a\left(l^g_t \mid \hat{l}^g_{t-1},l_{ctx}\right)\right), \hat{l}^g_t=\boldsymbol{g}_s\left(\hat{y}^g_t, \hat{l}^g_{t-1},l_{ctx}\right).
\end{align}
Here, $l_{ctx}$ denotes the group context, which is extracted from the I-latent $l^g_1$. Since the causal encoder concentrates spatial information into $l^g_1$, during compression we not only use the cached previous latent but also employ the  I-latent as the group-wide context for compression.

\begin{figure*}[t]
\centering
\includegraphics[width=1\linewidth]{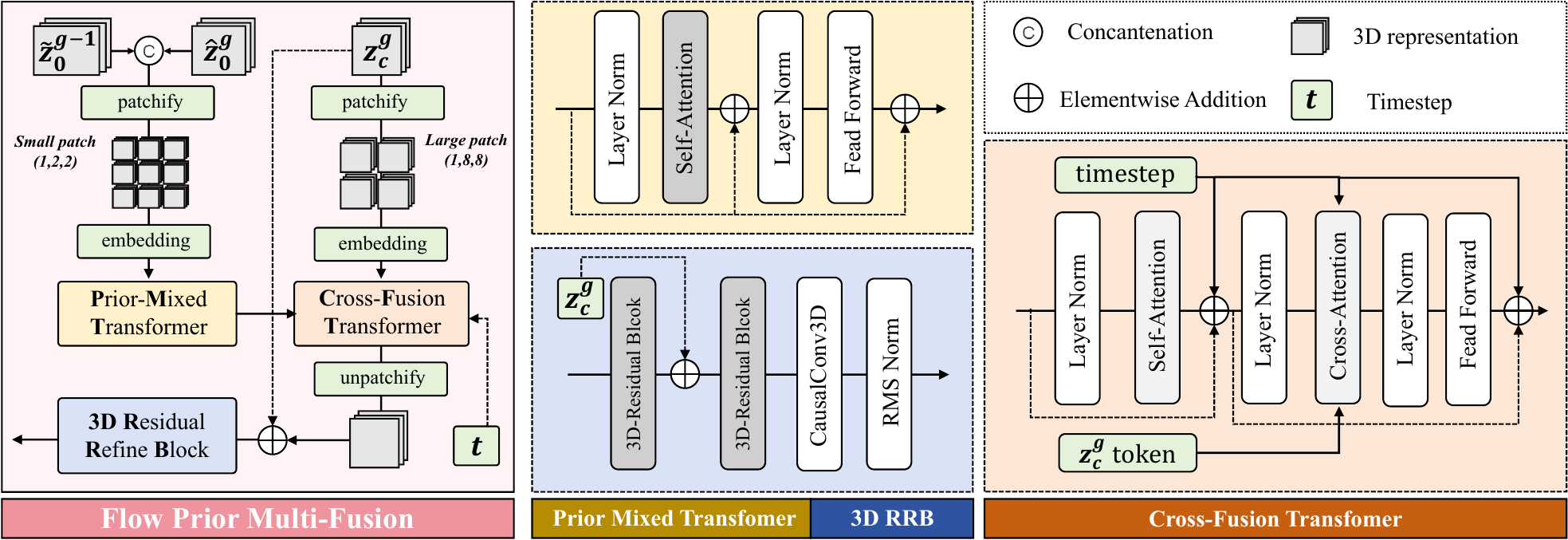}
\caption{Architecture of the Flow Prior Multi-Fusion (FPMF) module. By adopting patchify strategies with different granularities, FPMF achieves precise alignment between prior texture and structural features of compressed latent, facilitating more compact encoding and high-fidelity reconstruction.}
\label{fig:module}
\end{figure*}

Subsequently, we estimate $z^g_c=\{\hat{l}^g_t\}^{1+T/4}_{t=1}$ to determine the approximate position of the decoded compressed representation along the flow path:
\begin{align}
\lfloor t \rceil=\Gamma (z^g_c,z^g_0).
\end{align}

This information is encoded and transmitted to the decoder so that FlowDIT can treat the compressed representation as a state along the flow and perform flow-matching denoising:
\begin{align}
\hat{z}^g_0=\mathbf{FlowDIT}(z^g_c,\lfloor t \rceil)
\end{align}
For further refinement, the flow-prior reconstruction $\hat{z}^g_0$ is fed into the prior fusion network $\mathcal{F}(\cdot,\cdot,\cdot)$ together with the compressed representation $z^g_c$:
\begin{align}\tilde{z}^g_0=\mathcal{F}(\hat{z}^g_0,z^g_c,z_{ctx})
\end{align}

Here, $z_{ctx}$ is the cached prior context. In the key group, since no cache is available, we use $z^g_c$ as a substitute.

Through end-to-end training, this network implicitly fuses and aligns the flow representation domain and the compressed representation domain. Subsequently, $\tilde{z}^g_0$ is stored in the prior cache, serving as $z_{ctx}$ to assist reconstruction in P-groups.

Finally, at the decoder side, the 3D analysis decoder $\mathcal{D}(\cdot)$ decodes the reconstructed representation to the image domain:
$$
\hat{\boldsymbol{V}}_g=\mathcal{D}(\tilde{z}^g_0)
$$

\subsubsection{Contact Group to Group (CGG)}
\label{subsubsec:CGG}
Pre-trained video foundation flows cannot guarantee consistency across frame groups, and blindly increasing the number of frames within a group not only incurs substantial computational overhead but also limits the algorithm's adaptability to videos of different lengths. To address this, we design a non-intrusive caching mechanism to establish connections between groups, as shown in the \Cref{fig:pipeline}: the last frame of the I-Group enters the frame cache to serve as the first frame of the P-Group; simultaneously, the last frame of the reconstructed video $\hat{\boldsymbol{V}}_g$ from the I-Group is re-encoded into the latent cache for subsequent use.

Subsequently, when the compressor performs latent compression via (2), we use the cached $\tilde{l}^{g-1}_1$ from the latent cache to extract the group context $l_{ctx}$, and reconstruct the compressor's decoding output $z^g_c$:
$$
z^g_c=\{\tilde{l}^{g-1}_1\}+\{\hat{l}^g_t\}^{1+T/4}_{t=2}
$$
In this way, we encode the temporal context into the latent representation through the analysis encoder $\mathcal{E}(\cdot)$, but we neither encode nor transmit the key latent; instead, we substitute it with the cached latent. Grounded on the last decoded frame of the preceding group, this inter-frame spatiotemporal information guides the reconstruction of subsequent frames, thereby maintaining inter-group consistency. Meanwhile, the prior reconstruction result of the preceding group also enters the prior cache, to be retrieved as $z_{ctx}$ during prior fusion in subsequent groups. 

This design offers three advantages. First, it fully exploits inter-group context, further reducing the bitrate while maintaining reconstruction quality. Second, it avoids the artifacts—such as flickering and visual discontinuities—that would arise from inter-group independence in the reconstructed video. Third, it allows flexible adjustment of the ratio between P-Groups and I-Groups, enabling training-free bitrate selection within a certain range.

\subsection{Compress latent for flow-matching}
\label{subsec:CLF}
We aim to treat the decoded compressed representation as a state along the flow path, thereby non-intrusively learning the prior distribution pattern of the foundation flow-matching model. To this end, we need to compute the shortest distance from $z^g_c$ to the flow path. We assume that the compressed representation follows a normal distribution: $z^g_c\sim \mathcal{N}\left(z^g_0, \sigma^2\right)$. Given time $t$, the distribution along the flow path is $\mathbf{FL}(t)\sim \mathcal{N}\left((1-t)z^g_0, t^2\right)$. We seek to find the following time step $t^{\star}$:
$$
t^{\star}=\underset{t}{\arg \min }\, \mathbf{dist}\left[(1-t)*z^g_c, \mathbf{FL}(t)\right],t\in[0,1]
$$
Here, we also apply a corresponding scaling to $z^g_c$ to accommodate the mean shift caused during the flow matching process. $\mathbf{dist}$ denotes a certain distance metric. When we use the Wasserstein distance, the solution can be readily obtained (see the appendix for the proof):
$$
t^{\star}=\frac{\sigma}{1+\sigma}
$$
At this point, we can estimate the variance $\sigma^2=\|z^g_c-z^g_0\|_2$ based on the compressed representation $z^g_c$ and the input latent $z^g_0$, compute $t^{\star}$, and round it to obtain $\lfloor t \rceil=\mathbf{round}(t*1000)$. For video transmission tasks, the number of bits required to transmit the integer $\lfloor t \rceil$ is minimal and almost negligible.

After estimation, we treat $z^g_c$ as the flow state at time step $\lfloor t \rceil$, and use the flow-matching network for denoising:
\begin{align}
&\hat{z}^g_0=\mathbf{FlowDIT}(z^g_c,\lfloor t \rceil) \\
&=(1-\frac{\lfloor t \rceil}{1000})z^g_c-\mathbf{v}_{\theta}((1-\frac{\lfloor t \rceil}{1000})z^g_c,\lfloor t \rceil,ctx)
\end{align}
Here, $ctx$ is a fixed text semantic embedding. As shown in the \Cref{fig:patch}(b), our proposed estimation method avoids the absence of denoising and the over-smoothing caused by erroneous temporal estimation, thereby better learning the distribution pattern of the flow prior.

\subsection{Fusing Flow Prior for E2E training}
\label{subsec:FPMF}
The prediction of the flow-matching network cannot be directly used as the reconstruction result, for two main reasons. First, the error in time estimation and the discrepancy between the compressed representation distribution and the true flow path distribution would degrade reconstruction quality. Second, without accessing the gradients of the flow-matching network, the compressor cannot remove prior redundancy through end-to-end training. To address this, we design \textbf{F}low \textbf{P}rior \textbf{M}ulti \textbf{F}usion (FPMF), a Transformer-based prior correction and alignment network, whose architecture is shown in the \Cref{fig:module}.

\begin{figure}[t]
\centering
\includegraphics[width=1\linewidth]{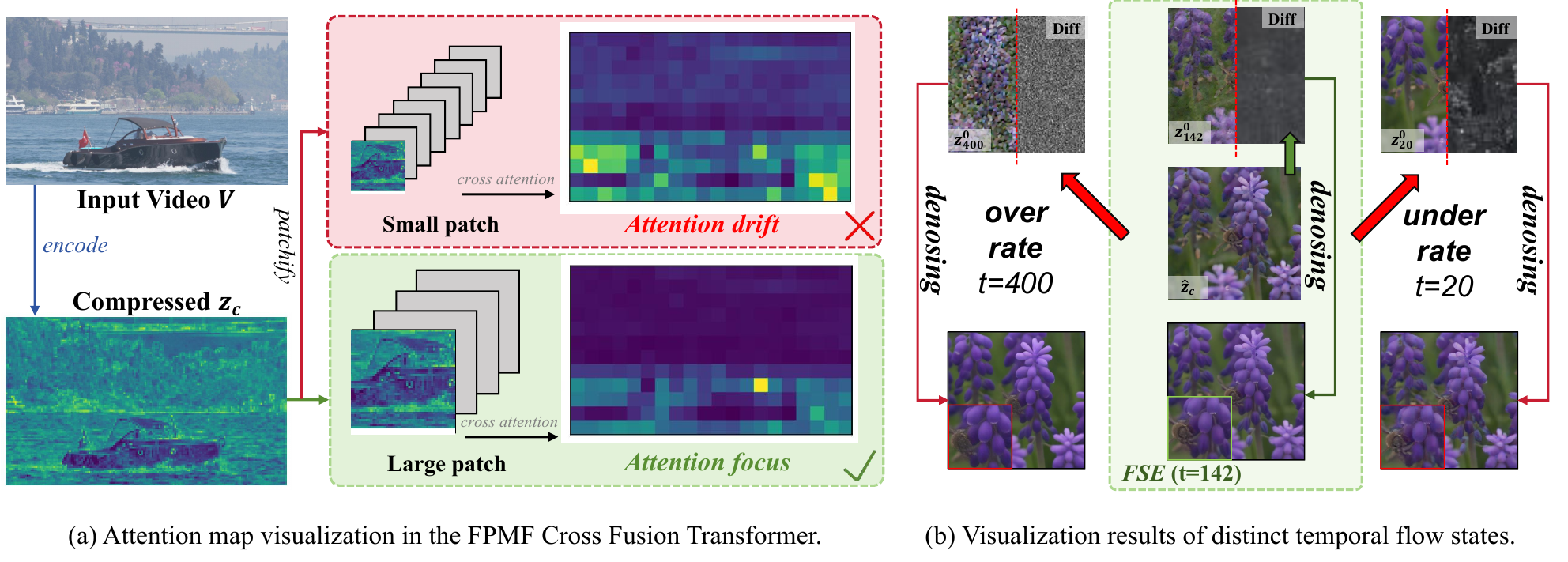}
\caption{Visualizations of the modules and selected methods. (a) Coarse-grained patchify drives compressed tokens to capture structural information, while fine-grained patchify distributes attention over latent space. (b) Overestimated or underestimated predictions cause the denoising network to either skip denoising entirely or render the image excessively smooth. Flow-State Estimation (FSE) precisely calculates the temporal state with high accuracy.
}
\label{fig:patch}
\end{figure}

The network takes as input the flow-prior reconstruction of the current latent group $\hat{z}^g_0$, the prior reconstruction of the preceding group $\hat{z}^{g-1}_0$, and the compressed representation $z^g_c$. The prior reconstructions are first concatenated and patchified into fine-grained small patches, since compared to the compressed representation, the prior reconstruction latents contain more high-frequency details, and we need a smaller patch size to attend to these textures as much as possible. In contrast, due to the ill-posed degradation and distortion introduced by compression, the details in $z^g_c$ are lost and unreliable. Therefore, we adopt a coarse-grained patchify strategy to learn the low-frequency non-local features of the compressed representation. The prior patches then pass through stacked ViT layers for mixing, and are subsequently fused and matched with $z^g_c$ in the Cross Fusion Transformer. As shown in \Cref{fig:patch} (a), when $\hat{z}^{g-1}_0$ is patchified into fine-grained patches, the attention is dispersed across texture-rich regions, which is unhelpful for reconstruction. Coarse-grained patches, on the other hand, avoid this issue and concentrate the attention on semantically rich regions such as contours and objects.

In fact, FPMF not only improves reconstruction performance but also further reduces the bitrate through a modeling approach guided by the principle of "compressing low-frequency structures while restoring high-frequency details via priors." We also verify this in the ablation experiments.

\subsubsection{Training Strategy}
\label{sec:Train}
$\mathtt{VoRTeC}$ is optimized end-to-end through rate loss and reconstruction loss across all modules. The rate loss constrains the bitrate of the compressor:
$$
\mathcal{L}_{rate}=R_{z^g_0}+\mathcal{L}_{\text {codebook }}
$$
Here, the rate $R_{z^g_0}$ is the aggregate of the rates of all frames: $R_{z^g_0}= \sum_{t=1}^{1+T/4} R_{l^g_t}$.

The reconstruction loss can be divided into three parts. The first part is the reconstruction loss in the latent domain $\mathcal{L}_{\mathbf{latent}}$, which constrains the compressor and the FPMF module:
$$
\mathcal{L}_{\mathbf{latent}}=\|z^g_0-z^g_c\|_2+\|\hat{z}_0^g-z_0^g\|_2
$$
The second part is the reconstruction loss in the image domain $\mathcal{L}_{\mathbf{rec}}$, which further improves the accuracy and perceptual quality of the reconstruction:
$$
\mathcal{L}_{\mathbf{rec}}=\|\hat{\boldsymbol{V}}_g-\boldsymbol{V}_g\|_2+\gamma_1\mathcal{L}_{\mathbf{lpips}}(\hat{\boldsymbol{V}}_g,\boldsymbol{V}_g)+\gamma_2\mathcal{L}_{\mathbf{dino}}(\hat{\boldsymbol{V}}_g,\boldsymbol{V}_g)
$$
The third part is the adversarial loss $\mathcal{L}_{\mathbf{adv}}(\hat{\boldsymbol{V}}_g,\boldsymbol{V}_g)$, which narrows the gap between the reconstructed video distribution and the real video distribution. In summary, the overall training loss is:
$$
\mathcal{L}=\lambda_1\mathcal{L}_{rate}+\lambda_2(\mathcal{L}_{\mathbf{latent}}+\mathcal{L}_{\mathbf{rec}})+\gamma_3\mathcal{L}_{\mathbf{adv}}
$$
Here, $\lambda_1$ and $\lambda_2$ are hyperparameters that control the rate-distortion trade-off, and $\gamma_1$, $\gamma_2$, $\gamma_3$ are fixed hyperparameters.

\section{Experiments}
\label{sec:experiments}
\subsection{Experimental Setup}
\label{subsec:setup}

\begin{figure*}[t]
\centering
\includegraphics[width=1\linewidth]{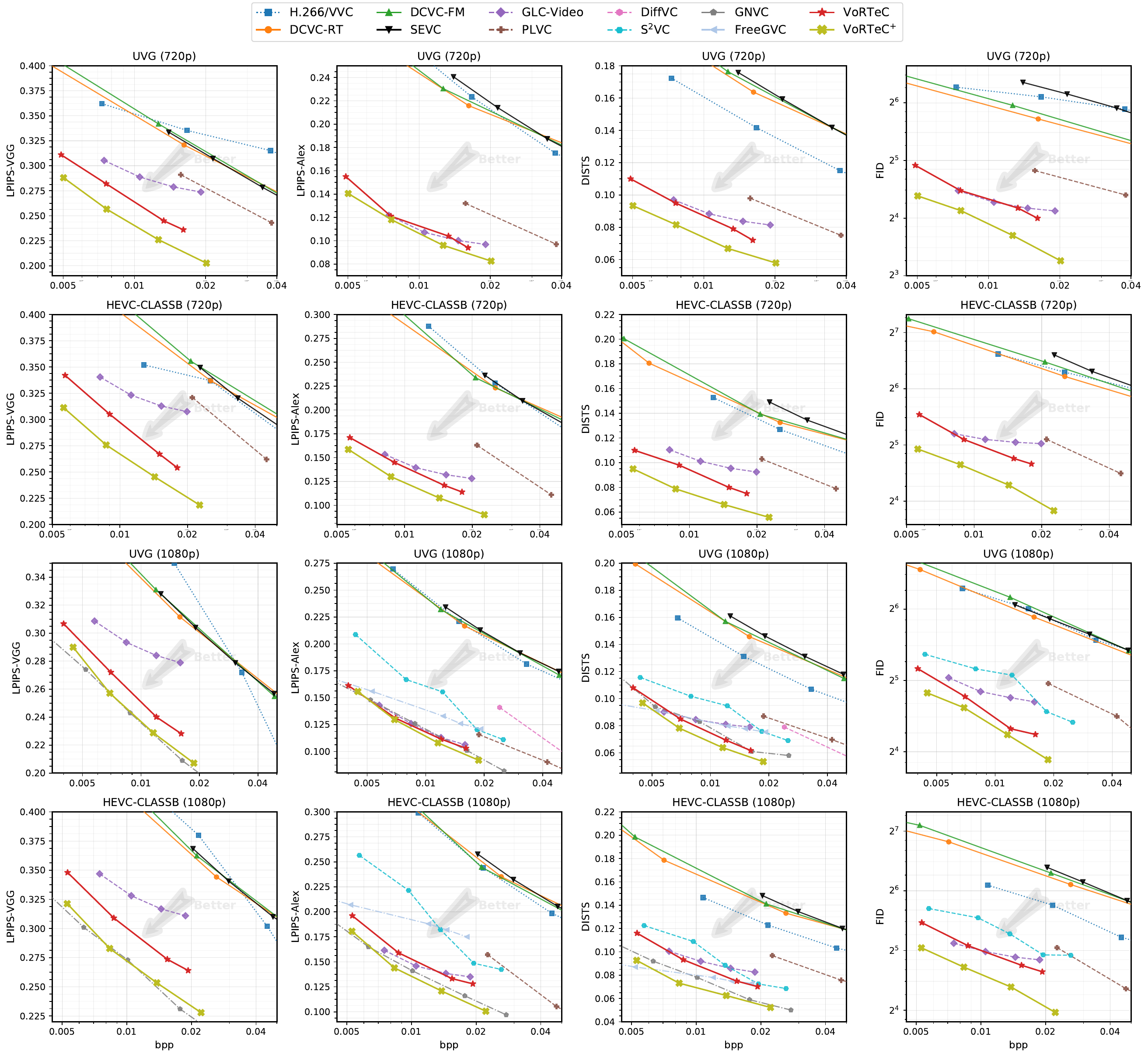}
\caption{Rate--distortion/perception curves on HEVC Class~B and UVG datasets at 720p and 1080p resolutions.}
\label{fig:rd_all}
\end{figure*}

\paragraph{Datasets.}
We evaluate our method on three standard video compression benchmarks: UVG~\citep{mercat2020uvg}, HEVC Class B,C \citep{sullivan2012hevc}, and MCL-JCV \citep{wang2016mcljcv}.
UVG, HEVC Class B and MCL-JCV contain $1920\times1080$ high-resolution natural videos.
HEVC Class C provides challenging test sequences at $832\times480$ resolution.
Following common practice in learned video compression~\citep{li2023dcvc,jia2025dcvcrt}, we test HEVC Class B and UVG under two protocols: (i) center-cropping to $1280\times720$ and encoding the full sequence; and (ii) evaluating the first 96 frames at the native $1920\times1080$ resolution.
All frames are processed in the RGB domain. For results on MCL-JCV and HEVC-Class C, please refer to the appendix.
\paragraph{Baselines.}
We compare against a comprehensive set of traditional, neural, and generative video codecs.
For traditional standards, we use VTM-17.0~\citep{bross2021vvc}, the reference implementation of H.266/VVCs.
For neural video codecs, we include DCVC-RT~\citep{jia2025dcvcrt}, DCVC-FM~\citep{li2024dcvcfm}, and SEVC~\citep{bian2025sevc}, representing the recent advances in conditional coding and spatially embedded compression.
For generative compression, we compare with GLC-Video~\citep{qi2025glcvideo}, PLVC~\citep{yang2022plvc}, DiffVC~\citep{ma2025diffvc}, S2VC~\citep{xue2025s2vc}, covering GAN-based and diffusion-based approaches. For certain diffusion-based baseline methods \citep{mao2025gnvcvd, li2026yoda} for which the source code is not publicly available, we report the performance metrics as presented in the respective original publications.

\paragraph{Metrics.}
We employ a suite of widely recognized quantitative metrics to assess the visual quality, including perceptual metrics such as LPIPS~\citep{zhang2018lpips} (vgg as default), DISTS~\citep{ding2020dists} and FID~\citep{heusel2017fid}.  Further, we utilize distortion metrics like PSNR. We additionally report FloLPIPS~\citep{danier2022flolpips} and $E_{warp}$~\citep{lai2018learning} to quantify temporal flickering artifacts. Meanwhile, we also report the BD-Rate \citep{bd} as the metric for rate-distortion performance.

\subsection{Main Results}
\label{subsec:results}
\begin{figure}[t]
\centering
\includegraphics[width=1\linewidth]{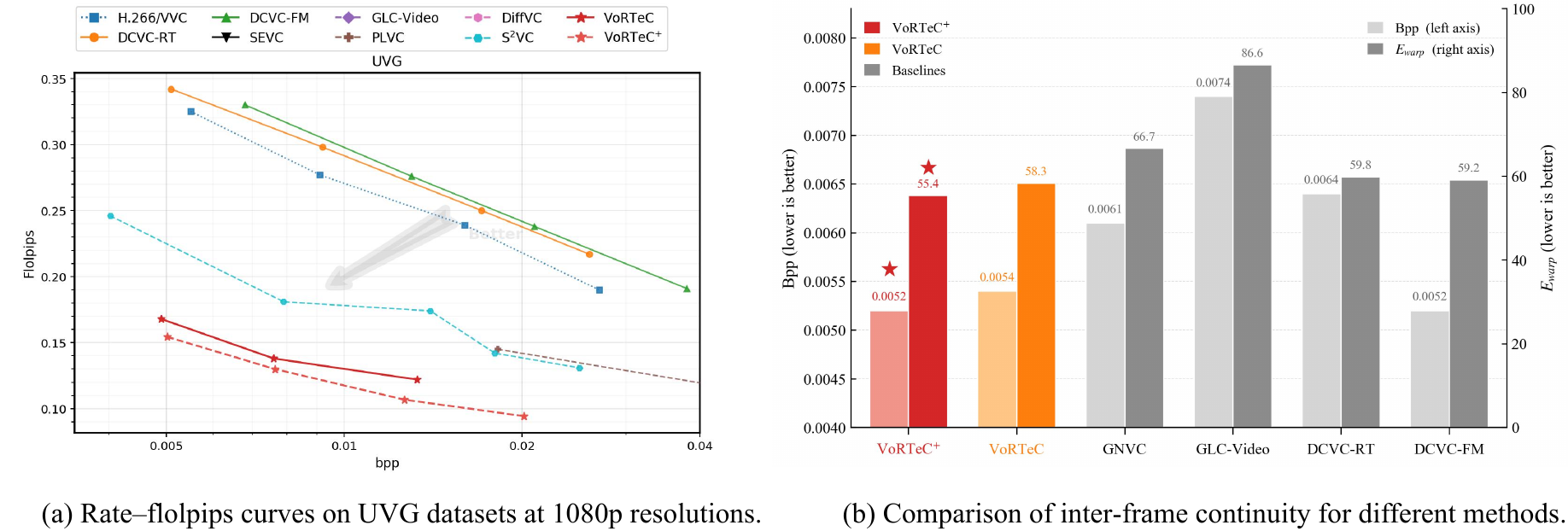}
\caption{Quantitative experiments for verifying inter‑frame continuity/consistency. (a) FloLPIPS metrics of various methods tested on the UVG dataset. Our proposed method consistently outperforms the baselines. (b) $E_{warp}$ metrics and corresponding bpp values of various methods tested on the HEVC‑ClassB dataset. $\mathtt{VoRTeC}$ achieves the minimal $E_{warp}$ even at the lowest bpp, indicating superior inter‑frame continuity.}
\label{fig:conti}
\end{figure}

Figure \ref{fig:rd_all} compares $\mathtt{VoRTeC}$ against baselines on HEVC-B and UVG. Traditional and distortion-optimized neural encoders (H.266/VVC, DCVC-RT, DCVC-FM~\cite{li2024dcvcfm}, SEVC) suffer from over-smoothing below $0.03$ bpp, reflected by high perceptual metric values. Generative baselines (GLC-Video, PLVC, DiffVC, S2VC) improve perceptual quality via adversarial or diffusion priors, yet remain inferior to $\mathtt{VoRTeC}$. At 720p, $\mathtt{VoRTeC}$ achieves consistent perceptual superiority without sacrificing distortion—it also outperforms the previous SOTA generative scheme GLC-Video in PSNR. At 1080p the advantage narrows but stays leading, which we attribute to the Wan2.1 1.3B base model being trained on 480p generation. $\mathtt{VoRTeC}$ shows consistent advantages on HEVC-ClassC (480p) as reported in the appendix, and qualitative visual comparisons in the supplementary materials further confirm its benefits. Benefiting from LoRA fine‑tuning, $\mathtt{VoRTeC}^{+}$ enables the prior distribution to adapt to high‑resolution compression degradation, yielding considerable performance gains. It substantially surpasses nearly all other generative video compression baselines in perceptual metrics and exhibits consistent superiority over GLC‑Video. We further present quantitative experiments on inter‑frame continuity of decoded frames in \Cref{fig:conti}. Our method outperforms other learning‑based video compression approaches across both metrics. Compared with GNVC \citep{mao2025gnvcvd}, which also adopts Wan2.1 as the backbone, $\mathtt{VoRTeC}$ achieves stronger inter‑frame consistency, which demonstrates the contribution of our scheme to extending the long‑range context of the model.

\paragraph{Complexity analysis.}
Table~\ref{tab:complexity} reports encoding and decoding throughput in frames per second (fps) across four resolutions on a single NVIDIA A100 GPU. VTM17 was tested on an Intel Xeon Gold 6330 CPU.
As can be observed, $\mathtt{VoRTeC}$ effectively addresses the speed-quality dilemma. Although $\mathtt{VoRTeC}$ builds upon a billion-parameter video flow-matching foundation model, it achieves a decoding speed of 3.93 frames per second at 1080p resolution. Specifically, it is $6.1\times$ faster than GNVC-VD~\citep{mao2025gnvcvd} (which also adopts Wan2.1), $4.1\times$ faster than YODA~\citep{li2026yoda}, and $3.1\times$ faster than S2VC, while maintaining comparable speed with DCVC-DC~\citep{li2023dcvc} and DCVC-FM.
The speed advantage of $\mathtt{VoRTeC}$ becomes more pronounced at lower resolutions. It achieves decoding frame rates of $105.25$, $32.31$, and $12.60$ fps at resolutions of 240p, 480p, 720p respectively. These results outperform all learned baselines, including DCVC-FM and GLC-Video.
\begin{figure*}[t]
\centering
\includegraphics[width=1\linewidth]{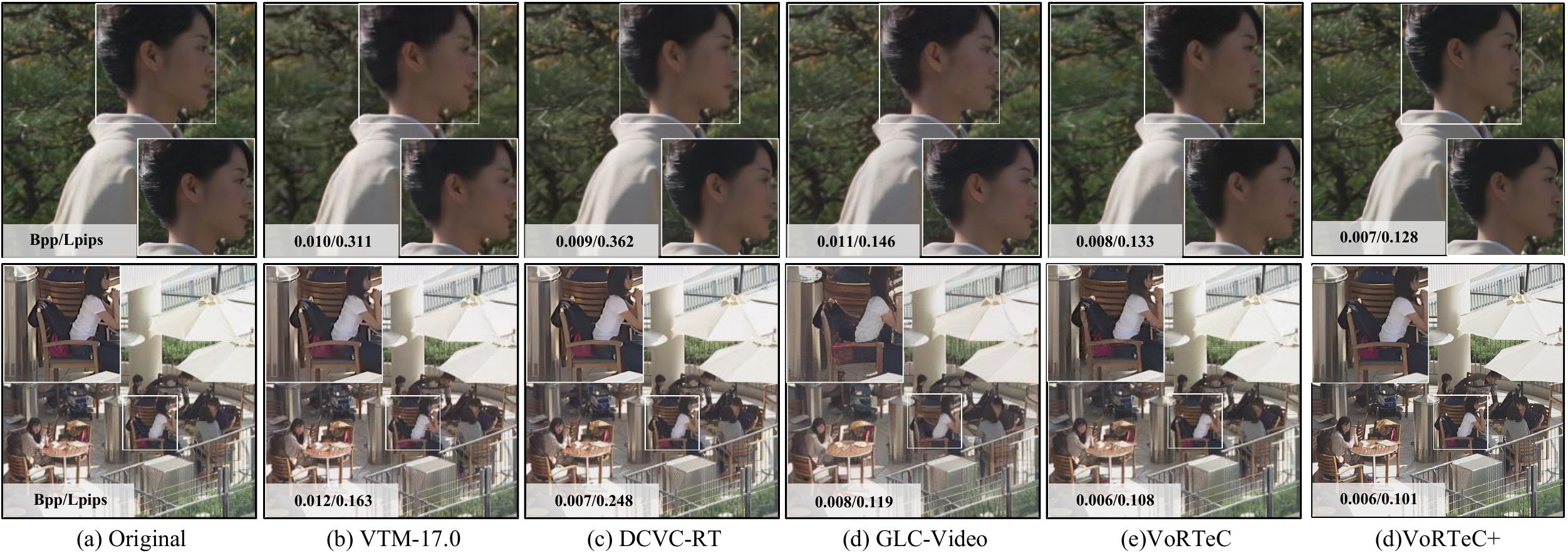}
\caption{Qualitative illustrations of various methods. \emph {Best viewed when zoomed in.}}
\label{fig:visual}
\end{figure*}
\subsection{Qualitative Results}

We present visual results of various state-of-the-art methods on the HEVC and UVG datasets, as shown in \cref{fig:visual}. Compared to MSE-optimized codecs (SEVC, DCVC series), $\mathtt{VoRTeC}$ reconstructs more realistic video frames, with clearer facial contours and details. After LoRA fine‑tuning, $\mathtt{VoRTeC}^{+}$ achieves improved reconstruction of details and textures, accompanied by a moderate reduction in bitrate.


\begin{table}
\centering
\caption{
Encoding and decoding throughput (fps) across four resolutions on a single A800 GPU. Higher is better.}
\label{tab:complexity}
\begin{tblr}{
  cells = {c},
  cell{1}{2} = {c=2}{},
  cell{1}{4} = {c=2}{},
  cell{1}{6} = {c=2}{},
  cell{1}{8} = {c=2}{},
  vline{2} = {-}{},
  hline{1} = {-}{0.08em},
  hline{2-3,12-13} = {-}{0.05em},
}
Resolution    & 416$\times$240 &          & 832$\times$480 &           & 1280$\times$720 &          & 1920$\times$1080 &          \\
Method        & Enc. Fps       & Dec. Fps & Enc. Fps       & Dec. Fps  & Enc. Fps        & Dec. Fps & Enc. Fps         & Dec. Fps \\
VTM-17    & 0.47           & 121.73   & 0.12           & 56.62     & 0.08            & 36.92    & 0.02             & 12.35    \\
DCVC-DC   & 36.24          & 43.88    & 15.63          & 18.71     & 7.47            & 9.36     & 3.43             & 4.36     \\
DCVC-FM   & 36.51          & 43.09    & 16.82          & 17.85     & 7.58            & 8.94     & 3.51             & 4.22     \\
GLC-Video & 49.39          & 57.22    & 28.60          & 19.92     & 13.25           & 8.82     & 6.36             & 4.19     \\
DiffVC    & 0.85           & 0.88     & 0.29           & 0.29      & 0.09            & 0.09     & 0.02             & 0.02     \\
GNVC-VD   & —~             & —~       &  $< 40.00$            & $< 7.75$     & 17.24           & 2.59     & 6.54             & 0.64     \\
FreeGVC   & —~             & —~       & $0.25 \sim1.38$~     & $0.32\sim1.25$ & —~              & —~       & —                & —        \\
S2VC      & —              & —~       & —~             & —~        & —~              & —~       & 6.60             & 1.27     \\
YODA      & —              & —~       & —~             & —~        & —~              & —~       & —                & 0.97     \\
$\mathtt{VoRTeC}$ (Ours)    & 84.46          & 105.25   & 20.34          & 32.31     & 8.76            & 12.60    & 3.84             & 3.93     
\end{tblr}
\end{table}

\subsection{Ablation Studies}

\begin{table}
\centering
\setlength{\extrarowheight}{0pt}
\addtolength{\extrarowheight}{\aboverulesep}
\addtolength{\extrarowheight}{\belowrulesep}
\setlength{\aboverulesep}{0pt}
\setlength{\belowrulesep}{0pt}
\begin{tabular}{c!{\vrule width \lightrulewidth}cc!{\vrule width \lightrulewidth}c!{\vrule width \lightrulewidth}cc} 
\toprule
\multicolumn{3}{c!{\vrule width \lightrulewidth}}{Module Ablation}                  & \multicolumn{3}{c}{Loss Ablation}                                                                                                       \\ 
\midrule
                & BD-rate\%(LPIPS) & BD-rate\%(FID) &                                                          & BD-rate\%(LPIPS)     & BD-rate\%(FID)        \\ 
\midrule
W/o FSE         & 44.31                            & 38.52                          & W/o $\mathcal{L}_{\mathbf{lpips}}$                                                & 40.31                                & 28.32                                 \\
W/o FPMF        & 72.59                            & 91.06                          & W/o $\mathcal{L}_{\mathbf{dino}}$                                             & 12.64                                & 21.37                                 \\
W/o Multi-patch & 16.24                            & 22.7                           & W/o $\mathcal{L}_{\mathbf{adv}}$                                              & 18.02                                & 37.15                                 \\ 
\hhline{~~~---}
W/o Prior-cache & 19.17                            & 23.83                          & \multicolumn{1}{c}{{\cellcolor[rgb]{0.753,0.753,0.753}}} & {\cellcolor[rgb]{0.753,0.753,0.753}} & {\cellcolor[rgb]{0.753,0.753,0.753}}  \\
W/ Lora Finetune        & -30.44                           & -45.32                         & \multicolumn{1}{c}{{\cellcolor[rgb]{0.753,0.753,0.753}}} & {\cellcolor[rgb]{0.753,0.753,0.753}} & {\cellcolor[rgb]{0.753,0.753,0.753}}  \\
\bottomrule
\end{tabular}
\caption{Ablation study on module components (top) and loss functions (bottom). BD-rate (\%) is computed relative to the full $\mathtt{VoRTeC}$ model on the UVG dataset. Lower is better.}
\label{tab:ablation}
\end{table}


\paragraph{Module Ablations}
 \Cref{tab:ablation} (top) presents ablation experiments for individual modules.
 
\emph{Flow State Estimation}: Instead of estimating the temporal state of the compressed representation, we use a fixed timestamp (t=300) during training. As can be seen, the perceptual quality degrades noticeably, indicating that the model struggles to learn the prior distribution pattern.

\emph{Prior Fusion}: We directly remove the FPMF module with no refinement on flow-prior reconstruction. As prior gradient information is unavailable in training, prior redundancy cannot be easily eliminated from the compressed representation, leading to a notable drop in perceptual quality.

\emph{Multi-patch}: For both the compressed representation and the prior reconstruction, we use the same mini patch size. As shown in \Cref{fig:patch}, this causes attention dispersion and leads to a certain degree of degradation in reconstruction performance.

\emph{Prior Cache}: We do not cache the prior representation $\tilde{z}^g_0$; instead, FPMF uniformly uses $z^g_c$ as the auxiliary input. This weakens the information flow between groups, leading to a certain degree of inter-frame instability and perceptual quality degradation.

\emph{Lora Fintune}: We fine‑tune FlowDiT using LoRA, namely $\mathtt{VoRTeC}^{+}$ mentioned in the main method. Fine‑tuning enables the denoising network to adapt to compression noise, which yields a noticeable improvement in decoding perceptual performance. Meanwhile, our scheme implicitly aligns compression noise with Gaussian noise; therefore, simple fine‑tuning allows the compression framework to achieve SOTA performance.
\paragraph{Loss Ablations}
\Cref{tab:ablation} (bottom) summarizes ablation experiments for different reconstruction loss functions. Both $\mathcal{L}_{\text{adv}}$ and $\mathcal{L}_{\text{dino}}$ effectively boost the visual realism of reconstructed images. Moreover, $\mathcal{L}_{\text{adv}}$ yields more prominent improvements in depth estimation accuracy for decoded frames. These observations reveal that enforcing feature-level consistency between reconstructed and original images within self-supervised model space helps preserve semantic and depth information of compressed images to the greatest extent.

\section{Conclusion}
\label{sec:conclusion}

This work presents $\mathtt{VoRTeC}$, a novel compression framework that enables one-step decoding using the pre-trained Wan2.1 video flow-matching foundation model. $\mathtt{VoRTeC}$ non-intrusively embeds spatio-temporal representations into the flow path and refines latent features via a multi-scale prior fusion module. Meanwhile, the proposed CGG scheme facilitates information interaction among frame groups and improves temporal continuity with negligible overhead.
Extensive experiments show that $\mathtt{VoRTeC}$ achieves an excellent rate-perception tradeoff and substantially accelerates decoding compared with previous diffusion-based methods, enabling real-time 32 FPS decoding at 480p. Benefiting from its flexible design, $\mathtt{VoRTeC}$ is compatible with flow-matching foundation models of various scales and architectures, providing a new paradigm for generative video compression.




\bibliographystyle{unsrtnat}
\bibliography{references} 

\newpage
\appendix

\section{Appendix}
\subsection{Mathematical Details}
\textbf{Derivation of $t^{\star}$ :}\\

We assume that the compression noise introduced by the compressor $\mathcal{C}(\cdot)$ based on the variational structure \citep{balle17} follows a Gaussian distribution with zero mean, i.e.:
\begin{align}
\mathcal{C}(z)=z_c\sim \mathcal{N}\left(\boldsymbol{z}, \sigma^2\right)
\end{align}
For a given time step $t$, we can compute the flow state at this time step according to the definition of normalizing flows \citep{lipman2023flowmatching}:
\begin{align}
\mathbf{FL}(t)=(1-t)z+t\boldsymbol{\epsilon}
\end{align}
Here, $\boldsymbol{\epsilon}$ follows the standard normal distribution $\mathcal{N}(0,\mathbf{I})$, and thus $\mathbf{FL}(t)$ also follows a normal distribution: $\mathbf{FL}(t) \sim \mathcal{N}((1-t)z,t^2\mathbf{I})$.

The problem discussed in the main text is to solve for $t^{\star}$:
\begin{align}
t^*=\underset{t}{\arg \min } \operatorname{dist}\left[(1-t) z_c, \mathbf{F L}(t)\right], t \in[0,1]
\end{align}
Here, $\mathbf{dist}(\cdot,\cdot)$ denotes the distance metric. We adopt the Wasserstein Distance, and the minimization problem is converted into finding the minimum value of the following expression:
\begin{align}
f(t)=(1-t)^2(z-z)^2+(|1-t|\sigma-|t|)^2 \\
=\left((1-t)\sigma-t \right)^2
\end{align}
It is easy to see that the minimum of $f(t)$ is $f(t^\star)=0$, which occurs at:
\begin{align}
t^{\star}=\frac{\sigma}{1+\sigma}
\end{align}
In our experiments, we approximate $\sigma$ with the sample variance to estimate $t^{\star}$.

\subsection{Additional Setting Details}
\subsubsection{Model Traning Details}
Our model uses Wan2.1 1.3B \footnote{https://huggingface.co/Wan-AI/Wan2.1-T2V-1.3B} as the foundation flow model, with the context model and entropy model based on modules from DCVC and ELIC. Training is divided into two stages. The first stage does not consider inter-group information flow, i.e., training is conducted in the I-Group mode. In this stage, we use a GOP size of 5 and train on the Vimeo-90k dataset \citep{viemo}. In the second stage, we expand the GOP size to 9 and set the number of frames within a meta-group to 17 (i.e., one I-Group and one P-Group), and fine-tune on the YouHQ dataset \citep{youhq}. The patch size is set to $256\times256$ throughout training. During training, we set the hyperparameters in the training objective as: $\gamma_1,\gamma_2,\gamma_3=2,0.2,0.1$. We adjust ${\lambda_1,\lambda_2}=\{1,6\},\{1,12\},\{1,25\},\{1,45\}$ to obtain different rate-distortion trade-offs.

Throughout all training stages, we employed AdamW \citep{loshchilov2017decoupled} as the optimizer. The learning rates were strategically set at \(1\times10^{-4}\)  for the initial phase and adjusted to \(5\times10^{-5}\) for the subsequent phase to facilitate finer parameter tuning. For $\mathtt{VoRTeC}^{+}$, we perform fine‑tuning using LoRA, where the rank of LoRA is set to 8. All experiments were conducted on an Nvidia A6000 GPU.

\subsubsection{Test setting Detail}

Original videos are stored in YUV420 format and converted to RGB according to the BT.709 standard. For codecs that require the input resolution to be a multiple of 64, zero-padding is applied before encoding; after decoding, the decoded frames are cropped to restore the original spatial dimensions. For 720p testing, we center-crop the original 1080p data to 720p.

\paragraph{Traditional Video Codecs}:

We evaluate VTM-17.0 \citep{bross2021vvc} in the RGB color space, adopting the settings from the DCVC series \citep{li2021dcvc}, converting the RGB input to 10-bit YUV444 format for internal codec processing. The configuration files used for each codec are as follows.

\begin{itemize}
\item VTM: \texttt{encoder\_lowdelay\_vtm.cfg}
\end{itemize}

The used parameters for each video as:
\begin{itemize}
\item \texttt{--c \{config file name\}}
\item \texttt{--InputFile=\{input video name\}}
\item \texttt{--InputBitDepth=10}
\item \texttt{--OutputBitDepth=10}
\item \texttt{--OutputBitDepthC=10}
\item \texttt{--InputChromaFormat=444}
\item \texttt{--FrameRate=\{frame rate\}}
\item \texttt{--DecodingRefreshType=2}
\item \texttt{--FramesToBeEncoded=\{frame number\}}
\item \texttt{--SourceWidth=\{width\}}
\item \texttt{--SourceHeight=\{height\}}
\item \texttt{--IntraPeriod=\{intra period\}}
\item \texttt{--QP=\{qp\}}
\item \texttt{--Level=6.2}
\item \texttt{--BitstreamFile=\{bitstream file name\}}
\end{itemize}

\paragraph{Neural Codecs}:

\textbf{DCVC-FM, DCVC-RT}: We use the officially released source code and pretrained weights \footnote{https://github.com/microsoft/DCVC}.

\textbf{GLC-Video}: We implement it using the open-source code and pretrained weights provided by GLC-Video \footnote{https://github.com/jzyustc/GLC}.

\textbf{SEVC}: We implement it using the open-source code and pretrained weights provided by SEVC \footnote{https://github.com/EsakaK/SEVC}.

\textbf{GNVC-VD, YODA, DiffVC, FreeGVC}: Since these works are not open-sourced, we use the original data reported in their respective papers while ensuring consistent experimental settings.

\textbf{S2VC}: We use the bitrate data directly provided by the original authors of S2VC.



For LPIPS, we utilized the \emph{lpips} library, while DISTS was implemented using \emph{DISTS\_pytorch}. FID metric were calculated using functions provided by \emph{torchmetrics.image}, with a feature size of 2048. Following the approach in \citep{mentzer2022hific}, during FID testing, images were partitioned. Specifically, from each $H\times W$ image, we initially extracted $\lfloor H / f\rfloor \cdot\lfloor W / f\rfloor$ non-overlapping $f\times f$ crops. Subsequently, the extraction origin was shifted by $f/2$ in both dimensions to obtain another $(\lfloor H / f\rfloor-1) \cdot (\lfloor W / f\rfloor-1)$ patches. A fixed value of $f = 256$ was used for all evaluations.

\subsection{Additional Experiment Results}
In this section, we present additional quantitative experimental results, including datasets and evaluation metrics not included in the main text. Furthermore, we conduct comprehensive visual comparisons between $\mathtt{VoRTeC}$ and the baseline methods.

\subsubsection{Quantitative Results}

\begin{figure*}[t!]
\centering
\includegraphics[width=1\linewidth]{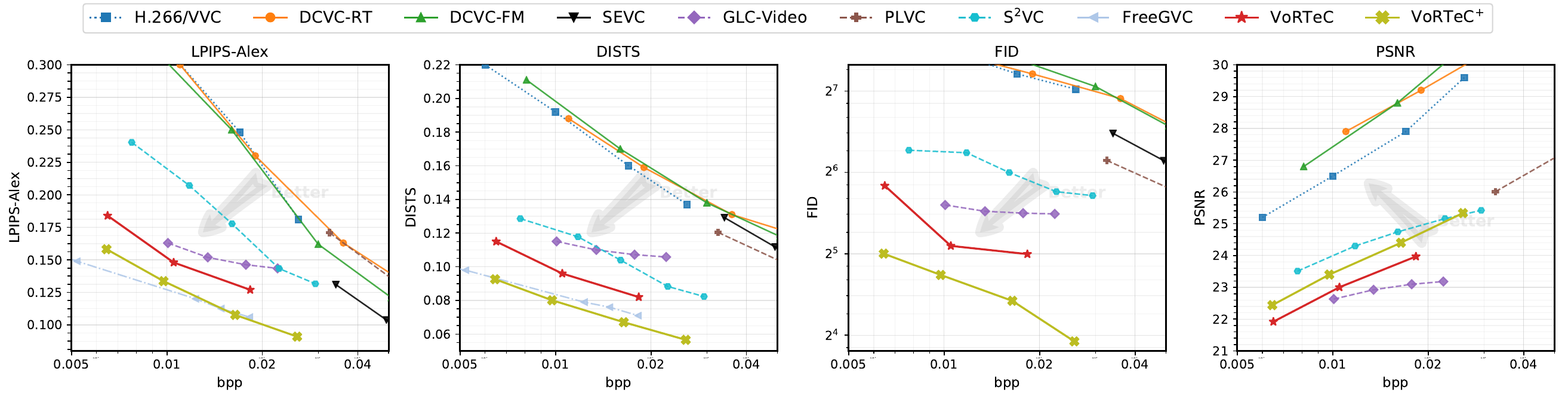}
\caption{Rate--distortion/perception curves on HEVC Class~C datasets.}
\label{fig:HEVCC}
\end{figure*}

\begin{figure*}[t!]
\centering
\includegraphics[width=1\linewidth]{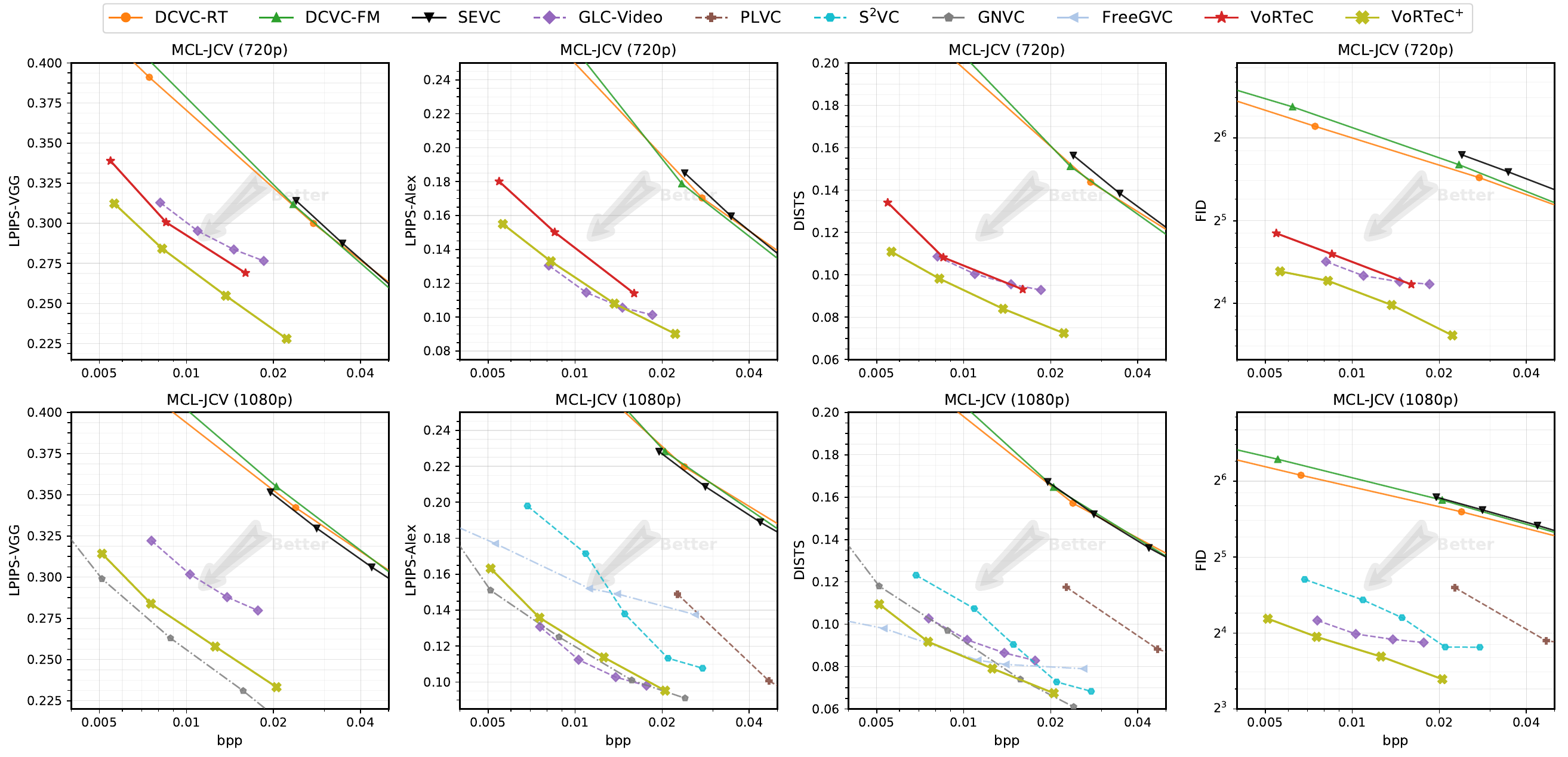}
\caption{Rate--perception curves on MCL-JCV datasets.}
\label{fig:MCLJCV}
\end{figure*}

\begin{figure*}[t!]
\centering
\includegraphics[width=1\linewidth]{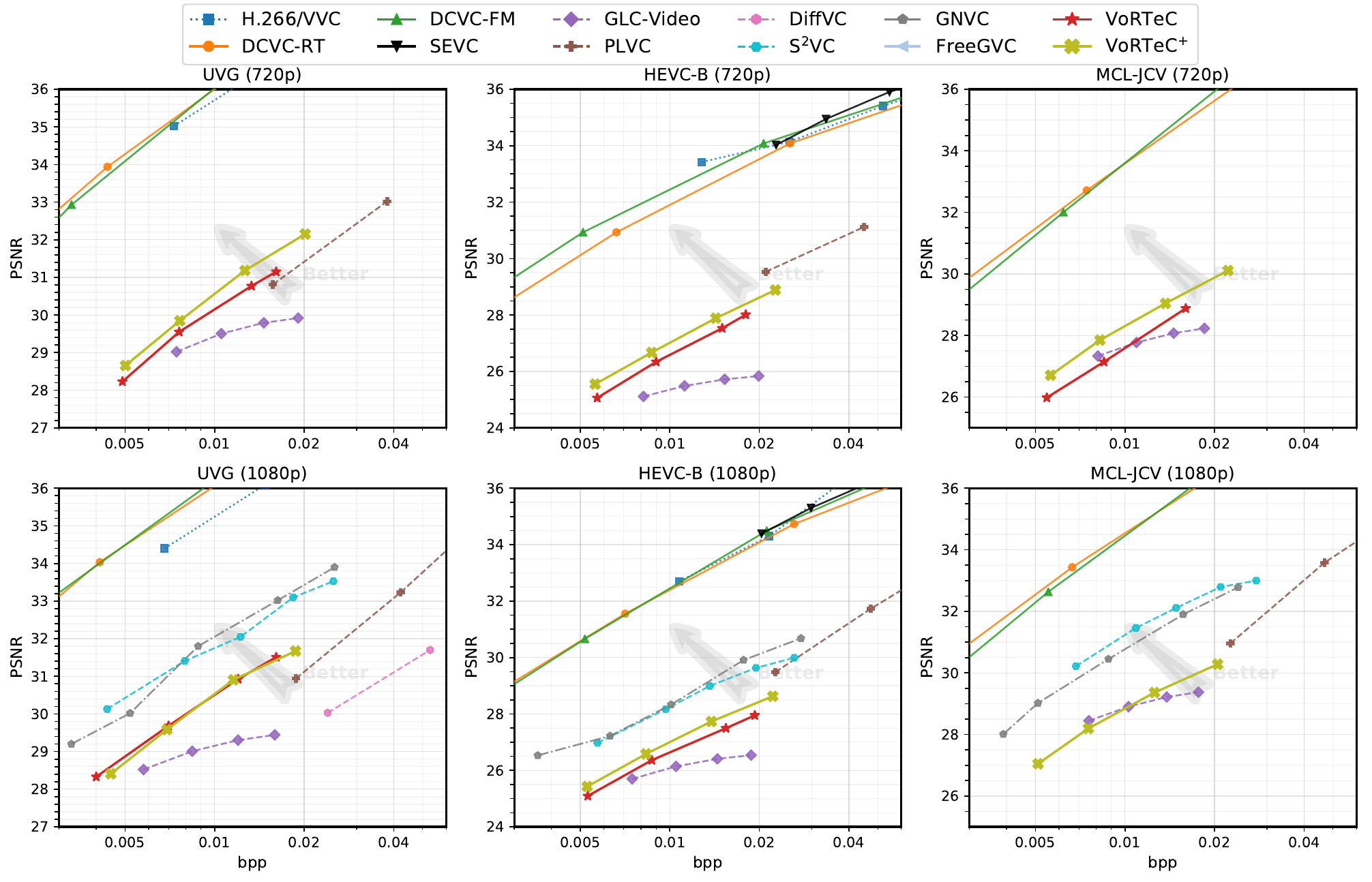}
\caption{Rate--distortion curves on HEVC Class~B, HEVC Class~C, UVG and MCL-JCV datasets.}
\label{fig:respsnr}
\end{figure*}

We first present the quantitative comparison results on the HEVC‑Class C and MCL‑JCV datasets, as illustrated in \Cref{fig:HEVCC} and \Cref{fig:MCLJCV}. Consistent with the observations and discussions in the main text, $\mathtt{VoRTeC}$ achieves superior performance on low‑resolution (480p) data, consistently outperforming generative methods across all metrics. On the MCL‑JCV dataset, our proposed solution also exhibits strong competitiveness. It consistently surpasses GLC‑Video on all metrics and delivers comparable performance to GNVC‑VD.

\Cref{fig:respsnr} reports comparative experiments for all remaining distortion metrics over the HEVC‑Class B, UVG, and MCL‑JCV datasets.
This figure validates the excellent compression capability of VoRTeC, which yields considerably higher PSNR than GLC‑Video.

In addition, we report the BD-Rate measured on all datasets (with $\mathtt{VoRTeC}$ as the anchor), as shown in Table 3. For some methods, the RD curves do not overlap with that of $\mathtt{VoRTeC}$, making it infeasible to compute the BD-Rate values.

\subsubsection{Ablation Result}

\begin{figure}[t]
\centering
\includegraphics[width=1\linewidth]{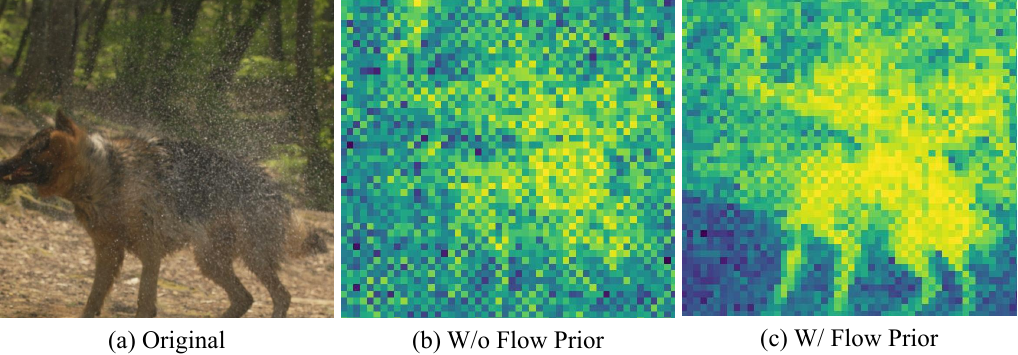}
\caption{Investigate the impact of the flow prior on learning the codec bit-rate distribution. Incorporating the flow prior enables the codec to learn more compact and rational bit allocation during end-to-end training.}
\label{fig:bits_allocation}
\end{figure}

\emph{Prior ablation}: \Cref{fig:bits_allocation} shows the ablation on bitrate allocation across modules. (a) shows the bitrate allocation map learned by the latent codec when the flow prior is removed, while (b) shows the bitrate allocation map after incorporating the flow prior. As can be seen, since $\mathtt{VoRTeC}$ removes the prior redundancy of the codec through end-to-end training, the bitrate is more accurately and reasonably allocated to specific objects rather than to flat background regions.

\begin{figure}[t]
\centering
\includegraphics[width=1\linewidth]{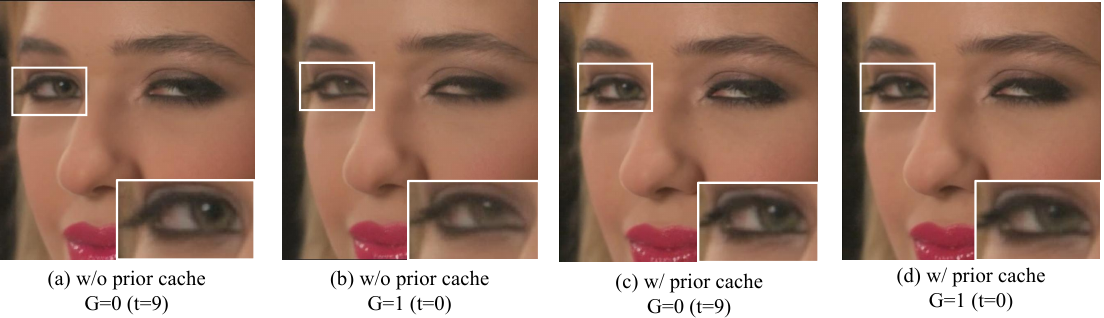}
\caption{The impact of prior caching on inter-group consistency is investigated. When prior caching is disabled, learning inter-group contextual information becomes difficult, resulting in abrupt changes to details including colors and contours.}
\label{fig:piror_ablation}
\end{figure}

\emph{Prior cahce ablation}: \Cref{fig:piror_ablation} demonstrates the contribution of the prior cache to inter-frame consistency. Without the prior cache, group transitions cause noticeable inconsistencies, manifesting as changes in texture, color, and shape (e.g., the color and size of the eyes), whereas the introduction of the prior cache effectively resolves this issue.

\begin{figure}[t]
\centering
\includegraphics[width=1\linewidth]{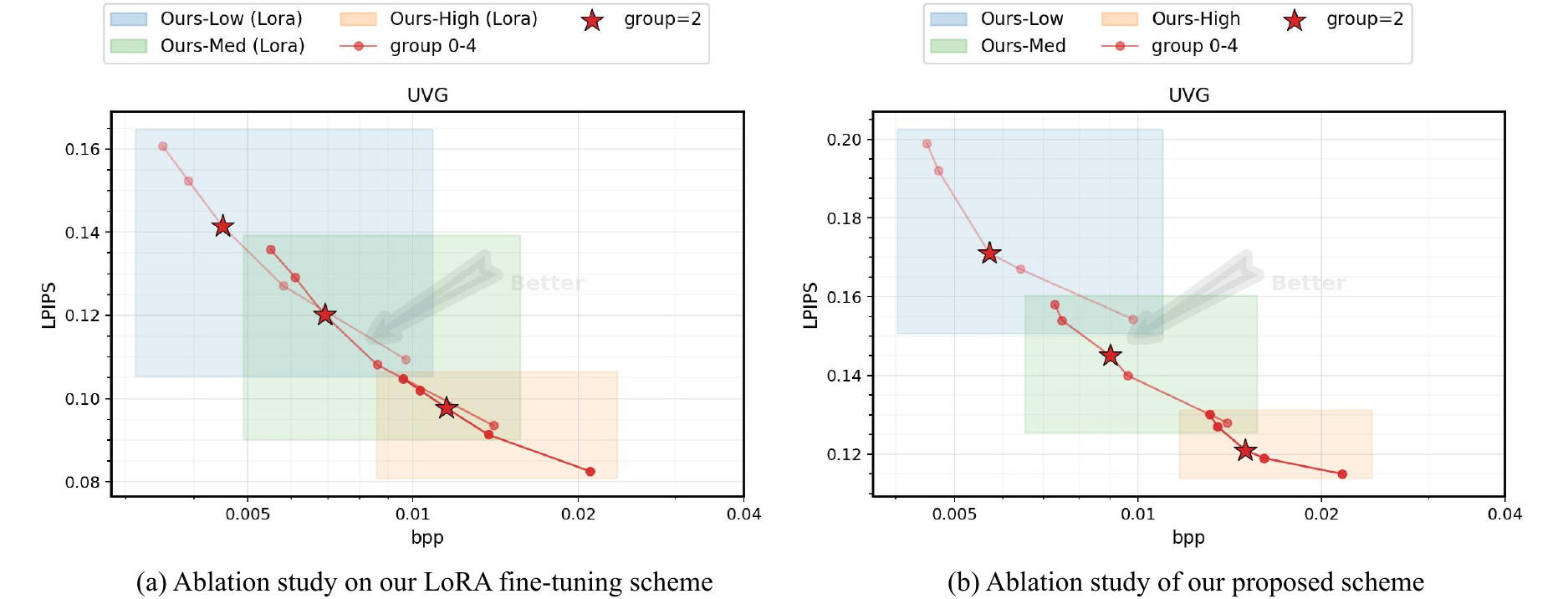}
\caption{We achieve a certain range of bit-rate scaling by adjusting the number of Pgroups within one Meta-group. Each color indicates the scalable bit-rate range supported by our single model, and the red star marks the number of Pgroups adopted in our main experiments. This further demonstrates the scalability and flexibility of $\mathtt{VoRTeC}$.}
\label{fig:metagroup}
\end{figure}

\emph{Meta Group ablation}: We conduct an ablation on the meta-group size. Since our designed CGG (Contact Group to Group) is scalable, we can increase the ratio of P-Groups to I-Groups to achieve bitrate adjustment within a certain range. In the main text, we set this ratio to 2, meaning that a meta-group contains 25 frames (9 + 8 + 8), consisting of one I-Group and two P-Groups. \Cref{fig:metagroup} demonstrates the scalability of our models at different bitrates. Reducing the number of P-Groups leads to an increase in bitrate, which in turn improves reconstruction quality. This indicates that our approach can flexibly cover a wider range of bitrates.

\begin{table}[t]
\centering
\small
\setlength{\tabcolsep}{5pt}
 \caption{BD-rate (\%) comparison on five datasets. All values are computed relative to $\mathtt{VoRTeC}$ (Ours). Positive values indicate that the corresponding method requires a higher bitrate than $\mathtt{VoRTeC}$ to achieve the same quality, \textit{i.e.}, $\mathtt{VoRTeC}$ is superior. ``--'' denotes unavailable results.}
\label{tab:bdrate}
\begin{tabular}{l *{5}{r}}
\toprule
Method & \multicolumn{1}{c}{LPIPS$_{\text{Alex}}$} & \multicolumn{1}{c}{LPIPS$_{\text{VGG}}$} & \multicolumn{1}{c}{DISTS} & \multicolumn{1}{c}{FID} & \multicolumn{1}{c}{PSNR} \\
\midrule
\multicolumn{6}{l}{\textit{(a) UVG 720p}} \\
\addlinespace[2pt]
GLC & -5.81 & 58.69 & 13.18 & -1.61 & 43.70 \\
DCVC-RT & {--} & 340.98 & {--} & {--} & \textbf{-88.45} \\
DCVC-FM & {--} & 344.47 & {--} & {--} & -85.06 \\
SEVC & {--} & 335.20 & {--} & {--} & {--} \\
VTM & 1302.42 & 925.28 & 899.59 & {--} & {--} \\
PLVC & 156.09 & 158.04 & 138.70 & 283.08 & 13.80 \\
$\mathtt{VoRTeC}$ & 0.00 & 0.00 & 0.00 & 0.00 & 0.00 \\
$\mathtt{VoRTeC}^{+}$ & \textbf{-12.32} & \textbf{-30.44} & \textbf{-37.21} & \textbf{-45.32} & -13.49 \\
\midrule
\multicolumn{6}{l}{\textit{(b) HEVC-B 720p}} \\
\addlinespace[2pt]
GLC & 15.15 & 61.58 & 45.95 & 24.46 & 75.61 \\
DCVC-RT & 1332.36 & 408.84 & 1217.26 & 1573.37 & -86.42 \\
DCVC-FM & 1187.45 & 426.88 & 1184.74 & {--} & \textbf{-89.22} \\
SEVC & 1171.48 & 383.14 & 1401.63 & {--} & {--} \\
VTM & 1062.20 & 381.97 & 763.14 & 2059.48 & {--} \\
PLVC & 189.02 & 184.93 & 180.85 & 130.14 & {--} \\
$\mathtt{VoRTeC}$ & 0.00 & 0.00 & 0.00 & 0.00 & 0.00 \\
$\mathtt{VoRTeC}^{+}$ & \textbf{-26.94} & \textbf{-33.94} & \textbf{-44.38} & \textbf{-51.20} & -16.57 \\
\midrule
\multicolumn{6}{l}{\textit{(c) HEVC Class C}} \\
\addlinespace[2pt]
GLC & 39.81 & 95.75 & 90.45 & 77.14 & 36.07 \\
SEVC & 103.04 & 201.26 & 570.04 & 1191.85 & {--} \\
S2VC & 113.87 & {--} & 80.88 & 244.79 & \textbf{-45.63} \\
PLVC & 301.00 & {--} & 476.05 & 925.16 & {--} \\
DCVC-FM & 239.47 & {--} & {--} & {--} & {--} \\
DCVC-RT & 319.31 & {--} & {--} & {--} & {--} \\
VTM & 293.43 & {--} & {--} & {--} & {--} \\
FreeGVC & \textbf{-41.37} & {--} & -44.21 & {--} & {--} \\
$\mathtt{VoRTeC}$ & 0.00 & 0.00 & 0.00 & 0.00 & 0.00 \\
$\mathtt{VoRTeC}^{+}$ & -30.89 & \textbf{-39.20} & \textbf{-47.56} & \textbf{-64.07} & -23.88 \\
\midrule
\multicolumn{6}{l}{\textit{(d) HEVC-B 1080p}} \\
\addlinespace[2pt]
GLC & -4.22 & 60.55 & 20.14 & 5.01 & 36.11 \\
DCVC-RT & 1195.73 & 452.70 & 1121.76 & 1651.50 & \textbf{-89.72} \\
DCVC-FM & 1028.40 & 454.92 & 1032.56 & {--} & -88.07 \\
SEVC & 1093.06 & 455.19 & 1072.35 & {--} & {--} \\
VTM & 976.40 & 364.72 & 605.10 & 572.67 & {--} \\
S2VC & 106.50 & {--} & 40.10 & 94.19 & -53.02 \\
PLVC & 119.76 & {--} & 202.20 & 108.71 & {--} \\
GNVC & -21.12 & \textbf{-34.58} & -30.47 & {--} & -44.96 \\
FreeGVC & 116.59 & {--} & -40.92 & {--} & {--} \\
$\mathtt{VoRTeC}$ & 0.00 & 0.00 & 0.00 & 0.00 & 0.00 \\
$\mathtt{VoRTeC}^{+}$ & \textbf{-25.73} & -34.40 & \textbf{-46.71} & \textbf{-46.52} & -15.49 \\
\midrule
\multicolumn{6}{l}{\textit{(e) UVG 1080p}} \\
\addlinespace[2pt]
GLC & 6.02 & 78.64 & 22.55 & 34.99 & 63.53 \\
DCVC-RT & {--} & 393.69 & {--} & {--} & \textbf{-88.64} \\
DCVC-FM & {--} & 388.62 & {--} & {--} & -87.01 \\
SEVC & {--} & 388.43 & {--} & {--} & {--} \\
VTM & 1430.52 & 350.12 & 858.28 & {--} & {--} \\
DiffVC & 273.79 & {--} & 178.37 & {--} & 203.22 \\
S2VC & 138.78 & {--} & 99.59 & 134.30 & -48.81 \\
PLVC & 79.10 & {--} & 236.45 & 309.67 & 49.87 \\
GNVC & 5.83 & \textbf{-23.88} & 1.63 & {--} & -40.55 \\
FreeGVC & 58.56 & {--} & 4.88 & {--} & {--} \\
$\mathtt{VoRTeC}$ & 0.00 & 0.00 & 0.00 & 0.00 & 0.00 \\
$\mathtt{VoRTeC}^{+}$ & \textbf{-5.36} & -21.22 & \textbf{-19.33} & \textbf{-17.88} & -1.80 \\
\bottomrule
\end{tabular}
\end{table}







\end{document}